\documentclass{svproc}
\usepackage{graphicx}
\usepackage{xcolor}
\usepackage{booktabs}
\usepackage{todonotes}
\usepackage{tabularx}
\newcommand{\xmark}{\ding{55}}%
\newcommand{\cmark}{\ding{51}}%
\usepackage{pifont}
\usepackage{url}

\begin{document}
\mainmatter              % start of a contribution
\title{Vision Based Machine Learning Algorithms for Out-of-Distribution Generalisation}
\titlerunning{Out-of-Distribution Generalisation}  % abbreviated title (for running head)
\author{Hamza Riaz\inst{1} \and Alan F. Smeaton\inst{2}}
\authorrunning{Hamza Riaz and Alan F. Smeaton} % abbreviated author list (for running head)
\tocauthor{Hamza Riaz, Alan F. Smeaton}
\institute{School of Computing\\
\and
Insight Centre for Data Analytics,\\
Dublin City University, 
Glasnevin, Dublin 9, Ireland.\\
\email{hamza.riaz2@mail.dcu.ie}
%\email{alan.smeaton@dcu.ie}
}

\maketitle              % typeset the title of the contribution

\begin{abstract}
There are many computer vision applications including object segmentation, classification, object detection, and reconstruction for which machine learning (ML)  shows state-of-the-art performance. Nowadays, we can build ML tools for such applications with real-world accuracy. However, each tool works well within the domain in which it has been trained and developed. Often, when we train a model on a dataset in one specific domain and test on another unseen domain known as an out of distribution (OOD) dataset, models or ML tools show a decrease in performance. For instance, when we train a simple classifier on real-world images and apply that model on the same classes but with a different domain like cartoons, paintings or sketches then the performance of ML tools disappoints. This presents serious challenges of domain generalisation (DG), domain adaptation (DA), and domain shifting. To enhance the power of ML tools, we can rebuild and retrain  models from scratch or we can perform transfer learning. In this paper, we present a comparison study between vision-based technologies for domain-specific and domain-generalised methods. In this research we highlight that simple convolutional neural network (CNN) based deep learning methods perform poorly when they have to tackle domain shifting. Experiments are conducted on two popular vision-based benchmarks, PACS and Office-Home. We introduce an implementation pipeline for  domain generalisation methods and conventional deep learning models. The outcome  confirms that CNN-based deep learning models show poor generalisation compare to other extensive methods.
% We would like to encourage you to list your keywords within
% the abstract section using the \keywords{...} command.
\end{abstract}
\section{Introduction}
The field of machine learning (ML) has created tremendous success stories by solving many complex problems like object classification, detection, segmentation and reconstruction in videos, natural language processing (NLP), medical image analysis, robotics, and many more. These developments in ML algorithms and databases, and the fusion of various fields of ML help researchers achieve high-level goals. The majority of current applications are built on what we call traditional ML where we usually have millions of example datapoints with labels to train a model under supervised learning (SL). 

Since $2011$, with the help of deep learning (DL) which is a sub-domain of ML that deals with various datasets to automatically extract features, scientists are now using DL to handle various supervised and unsupervised learning problems as described in \cite{b1}. Pure DL-based models are static systems, and have many problems including overfitting, they commonly need huge datasets, have data biases, and they do not have significant potential for generalisation and domain adaptation \cite{b2}.

Previous works illustrate that the majority of the time ML tools fail to generalise when processing out of distribution (OOD) data. The main reasons for wanting to design and analyse such generalised tools are applications like vision based autonomous systems, and medical imaging \cite{b2}. For example, when only a few conditions change during an inference process  in image processing such as  light variations,  shapes, locations, or the pose of objects, then models perform  poorly because they did not have interaction with similar variations during their training and thus did not learn how to perform under such unpredictable circumstances \cite{b3,b4,b5}. The work in \cite{b6} conveys  information about the collapse of ML tools for generalisation of OOD data which actually happens when ML models learn fake correlations instead of capturing real factors behind such variations in data. These fake correlations can be racial biases, texture statics, and object backgrounds. 

In the research literature, researchers have developed many methods to tackle domain generalisation. For instance, one of the first solutions that  was tried is about increasing the size of the training dataset with the same tasks but in different environments. The goal of the domain generalisation algorithm is to learn the invariances and features for all possible domain shifts. 

Before we go to the contributions this article, it is important to understand the difference between domain generalisation  and domain adaptation. When algorithms process samples of data from different distributions, ML algorithms suffer from a common problem called the domain shift. This  introduces two further major issues namely domain generalisation (DG) and domain adaption (DA). DG deals with the comparatively hard situations where several different but  related domains are given, and the purpose of a machine learning algorithm is to learn a model which could be generalised on unseen test data. The main goal of DG is learning a  representation of its training data that could have the potential to perform well in unseen domains by leveraging more source domains during training.  

The idea behind DA is different to DG in that it is to maximise the performance of algorithms or models on a given target domain using existing training source domains. The main difference between DA and DG is that DA has access to the target domain data which implies that it can see the data while DG cannot see anything from the target domain during training. This makes DG more challenging than DA but more realistic and more favourable in practical applications. There are many generalisation-related research topics or solutions such as domain adaptation, meta-learning, transfer learning, covariate shift, lifelong learning, and zero-shot learning.

\begin{figure}[htbp]
\includegraphics[width=1\textwidth]{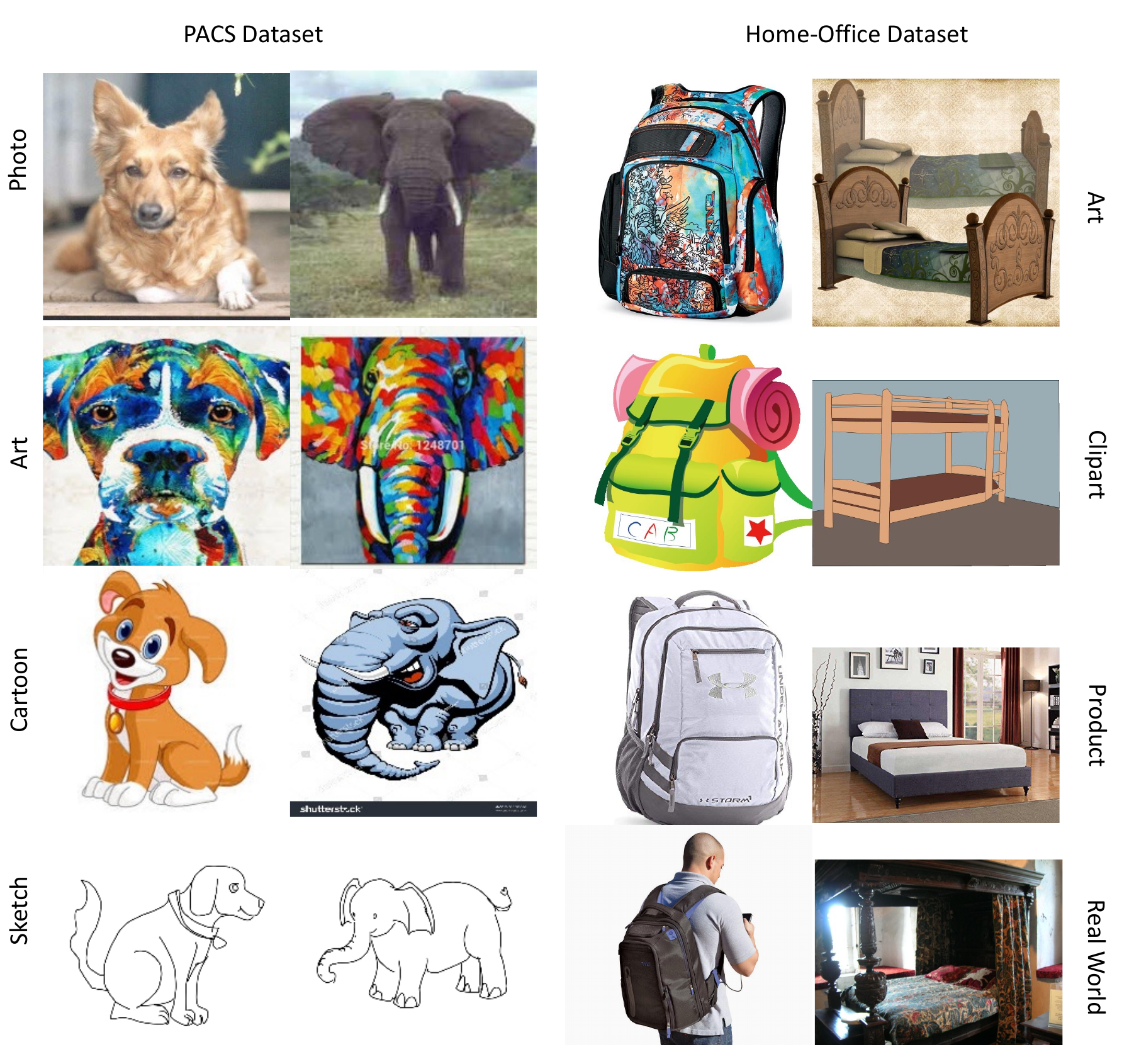}
\caption{Sample images of the same classes across all  domains in the PACS and Home-Office datasets.}
\label{fig1}
\end{figure}

This article addresses a direct comparison study between domain-specific and domain generalised methods with respect to vision-based applications, especially classification. To achieve our goal, we have implemented a pipeline with 9 well-known domain generalised algorithms and 7 domain-specific models. The comparison study is conducted on two popular benchmarks namely PACS and Office-Home, from which some sample images are shown in Figure~\ref{fig1}. We also trained and tested 16 models by using fine-tuning. Our research  shows the learning curves of methods for both benchmarks. The result section considers accuracy as a measure of generalisation for supervised learning benchmarks.

The rest of article is structured as follows: section 2 covers some  related work. Section 3 introduces brief details about the algorithms we use. Section 4 is about the study of benchmarks. Section 5 describes our  proposed method including the experiments and formulation of DG. Results and discussion are presented in Section 6 while Section 7 presents conclusions and future work.

\section{Related Work}

To handle domain generalisation, few methods have been proposed to select hyperparameters so that a model can maximise the performance for OOD \cite{b8}. The parameters update with the rest to a function which calculates relatedness between the different domains. Similarly, in \cite{b9} the authors illustrate  ways to select models based on an algorithm-specific regularisation. These previous articles worked for a specific kind of problem only, although the scope for this article is to analyse vision-based domain generalisation for various  up-to-date methods. 

In the literature we  find that domain generalisation has many available algorithms which can be classified into data manipulation, representation learning and strategy learning algorithms \cite{b7}. One piece of information which we can extract from \cite{b7} on data generation and adversarial training is also a kind of common way to optimise ML tools for OOD. In this regard, the well known ImagNet challenge \cite{b10} also updated for different kinds of OOD where authors created new benchmarks and then tried to solve them with new approaches \cite{b11,b12,b13}. Nevertheless, these methods did not consider solving  domain generalisation in vision-based applications with the most up-to-date approaches as explained in Domainbed \cite{b6}. Furthermore, \cite{b11,b12,b13} have variations in ImageNet challenge to solve domain generalisation paradigm even though they tried few domain specific methods, on other hand, totally missing out domain generalisation frameworks. Table~\ref{tab1} provides a clear picture in this regard.

%\Reviewer{Reviewer:2 Please point out not only the studied technologies/methods in the related work section but be clear as to how the previous work being described relates to the author’s own. Authors should critique the existing work - Where is it strong where is it weak? What are the unreasonable/undesirable assumptions?}

\begin{table}[htbp]

\caption{Comparison with previous articles for domain generalisation and domain specific methods where \cmark indicates whether an article has the details and \xmark means article is missing that point}
%\centering

\begin{tabularx}{\textwidth}{X*{5}{>{\centering\arraybackslash}X}} 
%\begin{tabularx}{\textwidth}{bsssss}
%\begin{tabular}{|c|c|c|c|c|c|{5cm}}
%\begin{tabular}{rlrrrr{5cm}}

\hline 
%\textbf{Table}&\multicolumn{3}{|c|}{\textbf{Table Column Head}} \\
 Articles & Model Selection & Training Framework Selection & Multiple Datasets & Domain Generalisation Frameworks & Domain Specific Frameworks \\ 
%\textbf{Articles}& \textbf{\textit{Model Selection}}& \textbf{\textit{Training Framework Selection}}& \textbf{\textit{Multiple Datasets}}& \textbf{\textit{Domain Generalisation Frameworks}}& \textbf{\textit{Domain Specific Frameworks}} \\
\hline
Gulrajani and Lopez-Paz \cite{b6} & \cmark & \cmark & \cmark & \cmark& \xmark  \\
Wang {\em et al.} \cite{b7} & \cmark & \cmark & \cmark & \cmark& \xmark \\
Hendrycks {\em et al.} \cite{b11} & \cmark & \xmark & \cmark & \xmark& \xmark \\
Hendrycks and Dietterich \cite{b12} & \cmark & \xmark & \xmark &\xmark & \cmark \\
Hendrycks {\em et al.} \cite{b13} & \cmark & \xmark & \xmark &\xmark & \cmark \\
Ours & \cmark & \cmark & \cmark & \cmark& \cmark \\
\hline
%\multicolumn{6}{l}{}

\label{tab1}

\end{tabularx} 

\end{table}

The most related state-of-the-art research papers are \cite{b6} and \cite{b7}. In \cite{b6}, the authors implemented a framework named Domainbed which has support for various domain generalised methods to analyse vision-based domain generalisation. Moreover, Domainbed has variety in model selection, training schemes and hyper parameters using Resnet as backbone model for domain generalisation frameworks. The anaylsis provided by authors in those papers were insightful. However, \cite{b6} does not has information about domain specific methods and does not provide training and inference analysis for domain specific models. 

Similarly, \cite{b7} also conducted the same kind of study but improve the limitations of Domainbed in implementation and coding flexibility. However, neither of these works  discussed the effect or performance of  traditional deep learning methods for OOD generalisation which is also the scope of our research. In this paper, we   perform an analysis of typical deep learning methods and then compare their performance with domain generalised methods. Therefore, table~\ref{tab1} describes the gaps in other works and improvements in terms of contributions for our article. 

%\Reviewer{Reviewer:2 The explanation of previous similar studies has been well presented, but at the end of the related work section, an important point must be added in the form of an explanation of what the limitations of the similar research areas, which can also be presented in the table, then what is the author's proposed to overcome this gap.}
\section{Implemented Algorithms}
Our paper uses an implementation of 16 algorithms in total, including 9 domain generalised and 7 conventional deep learning. This section briefly enumerates each of them.
\subsection{Up-to-Date Algorithms for Vision-Based Generalisation}
\begin{itemize}
\item Empirical risk minimisation (ERM) is a simple method which actually minimises the total sum of all errors in the given domains \cite{b6}.
\item Group distributionally robust optimisation (DRO) \cite{b14} also performs ERM but gives more focus to the domains with larger errors. We can say that it is an extension of simple ERM.
\item Inter-domain mixup (Mixup) \cite{b15,b16} uses ERM on linear interpolations of data in domains. 
\item Domain adversarial neural networks (DANN) \cite{b17} explore features with distribution matching in the external domains.
\item Class-conditional DANN also known as C-DANN \cite{b18}, is the extension of DANN and instead of matching features in the data distributions, it matches conditional distributions across the domains and their respective data labels.
\item Deep CORAL \cite{b19} utilises the matching between the mean and covariance of features across the distributions.
\item Maximum mean discrepancy (MMD) \cite{b20} measure the alignment of distributions across all the domains and use adversarial feature learning to match aligned distributions.
\item Invariant risk minimisation (IRM) \cite{b21} learns a linear classifier on top of representation matching techniques.
\end{itemize}
\subsection{Conventional Deep Learning Algorithms for Vision-Based Applications}
We implemented 7 well-known deep learning networks including AlexNet, VGGNet16, ResNet18, ResNet50, InceptionV3, DenseNet, and SqueezeNet, each of which are covered in an overview article \cite{b22}. These networks are popular in computer vision applications, therefore further details on them may not be needed.

The most related state-of-the-art research papers are \cite{b6} and \cite{b7}. In \cite{b6}, the authors implemented a framework named Domainbed which has support for various domain generalised methods to analyse vision-based domain generalisation. Similarly, \cite{b7} also conducted the same kind of study but improve the limitations of Domainbed in implementation and coding flexibility. However, neither of these works  discussed the effect and performance of  traditional deep learning methods for OOD generalisation which is also the scope of our research. In this paper, we  will perform an analysis of typical deep learning methods and then compare their performance with domain generalised methods. 
\begin{table*}[htbp]
\caption{Benchmarks used in supervised learning}
%\centering
\begin{center}

\begin{tabular}{lcccp{5cm}}

\hline 
%\textbf{Table}&\multicolumn{3}{|c|}{\textbf{Table Column Head}} \\
\textbf{Datasets} & \textbf{\textit{Domains}}& \textbf{\textit{Classes}}& \textbf{\textit{Samples}}& \textbf{\textit{Descriptions}} \\
\hline
Office-Caltech & 4 & 10 & 2,533 & Caltech, Amazon, Webcam, DSLR  \\
Office-31 & 3 & 32 & 4,110 & Amazon, Webcam, DSLR \\
PACS & 4 & 7 & 9,991 & Art, Cartoon, Photos, Sketches \\
VLCS & 4 & 5 & 10,729 & Caltech101, LabelMe, SUN09, VOC2007 \\
Office-Home & 4 & 65 & 15,588 & Art, Clipart, Product, Real World \\
Terra Incognita & 4 & 10 & 24,788 & Wild animal images recoded at four different locations L100, L38, L43, L46 \\
Rotated MNIST & 6 & 10 & 70,000 & Rotated Hand written Digits \\
DomainNet & 6 & 345 & 586,575 & Clipart, Infograph, Painting, Quickdraw, Real, Sketch \\
\hline
\multicolumn{5}{l}{}
\end{tabular}
\label{tab2}
\end{center}
\end{table*}
\section{Benchmarks Used}

To carry out experiments on domain generalisation, we consider only vision-based benchmarks for supervised learning. We restrict this work to vision-based benchmarks in order to narrow the scope because if we were to work with benchmark datasets across multiple domains covering vision, robotics, language processing, etc. then our results would have an overhanging question of whether results would have had as much to do with the domains chosen. 

Table~\ref{tab2} presents a summary of some of the open benchmarks used in the literature for evaluating supervised learning. For our initial experiments, we use two of these, PACS and Office-Home benchmarks. In the case of reinforcement learning, RoboSuite, DMC-Remastered, DMC-GB, DCS, KitchenShift, NaturalEnvs MuJoCo, CausalWorld, RLBench, Meta-world and many others are commonly used benchmarks as described in \cite{b23}.

Of the two benchmarks we use, one is a relatively simple dataset (PACS) with 4 different domains including images presented as Art, Cartoon, Photos, and Sketches. Each domain has 7 classes and there are 9,991 samples in total. The second benchmark  is Office-Home, also consisting of images. This also has 4 domains namely Art, Clipart, Product, and Real World with 65 classes in each domain. The main idea behind choosing the first  is to work on  a benchmark which could have comparatively less complex classification tasks so that we can observe the behaviours of domain-specific and domain generic models. We select Office-Home as a second benchmark because of the higher number of classes (supervised tasks) which adds  complexity into the tasks for each domain.

\section{Experimental Methods}
\begin{figure}[hbt!]
\centering
\includegraphics[width=0.8\textwidth]{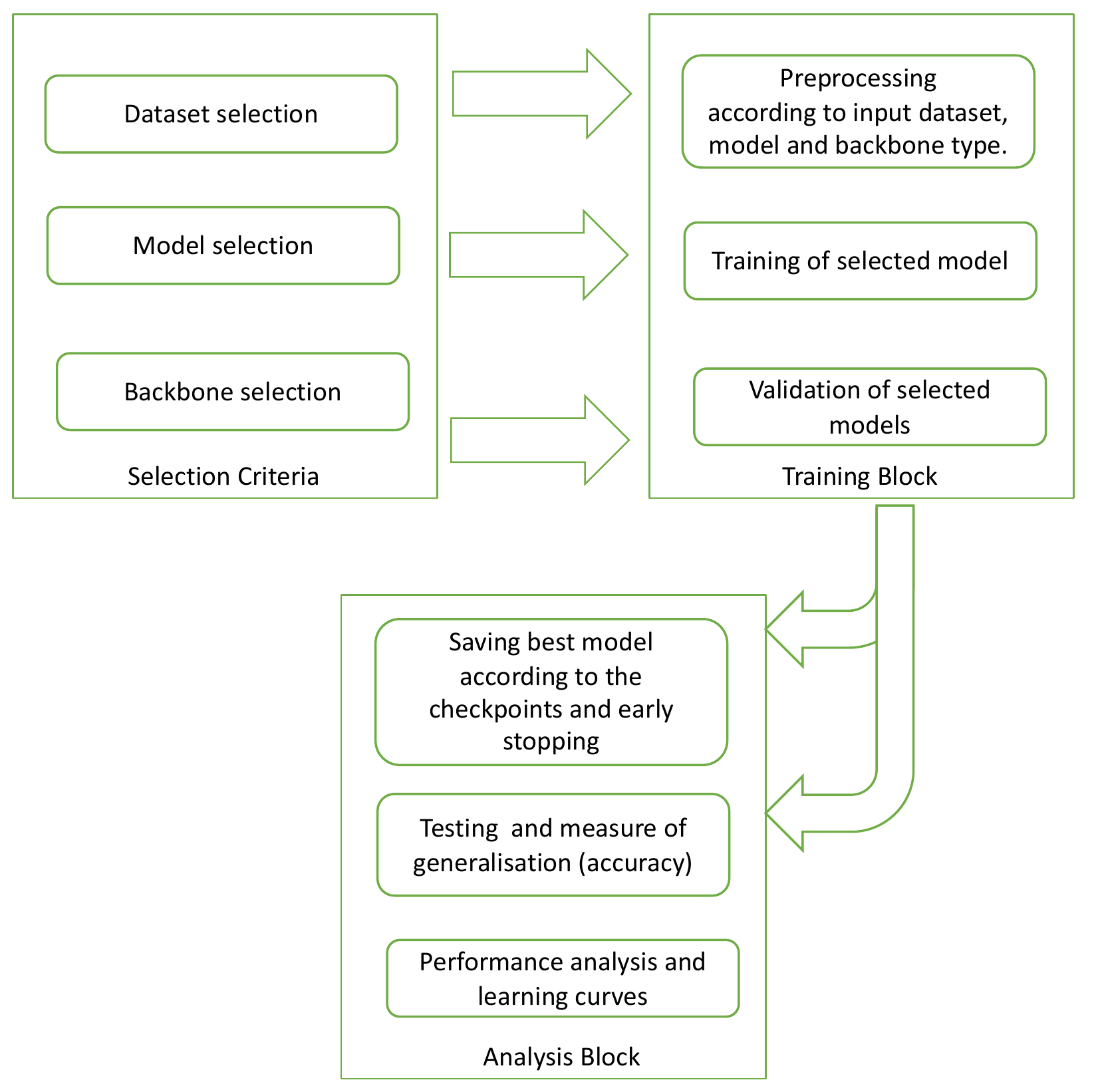}
\caption{Overview diagram summarising our work}
\label{fig2}
\end{figure}
To investigate our research questions we create a pipeline consisting of both types of algorithms and Figure~\ref{fig2} provides an overview of our approach. Figure~\ref{fig2} contains three blocks including selection criteria, training and analysis. In the first block, we  choose the dataset or benchmark for which we want to try the proposed pipeline. This block also includes the model selection step in which we identify which type of leaning method our pipeline will use, either  conventional deep learning like VGGNet, ResNet or one of the more recent vision-based  domain generalised methods like ERM, DROP, etc. The training block is the second block and includes data pre-processing according to selected conditions, training and validation of models. The third block,  analysis, saves the best model according to checkpoints and early stopping. It also measures the generalisation in the form of accuracy and loss metrics and compares the performance by computing learning curves.

\subsection{Formulation of DG and Experiments}
\begin{figure}[htbp]
\includegraphics[width=\textwidth]{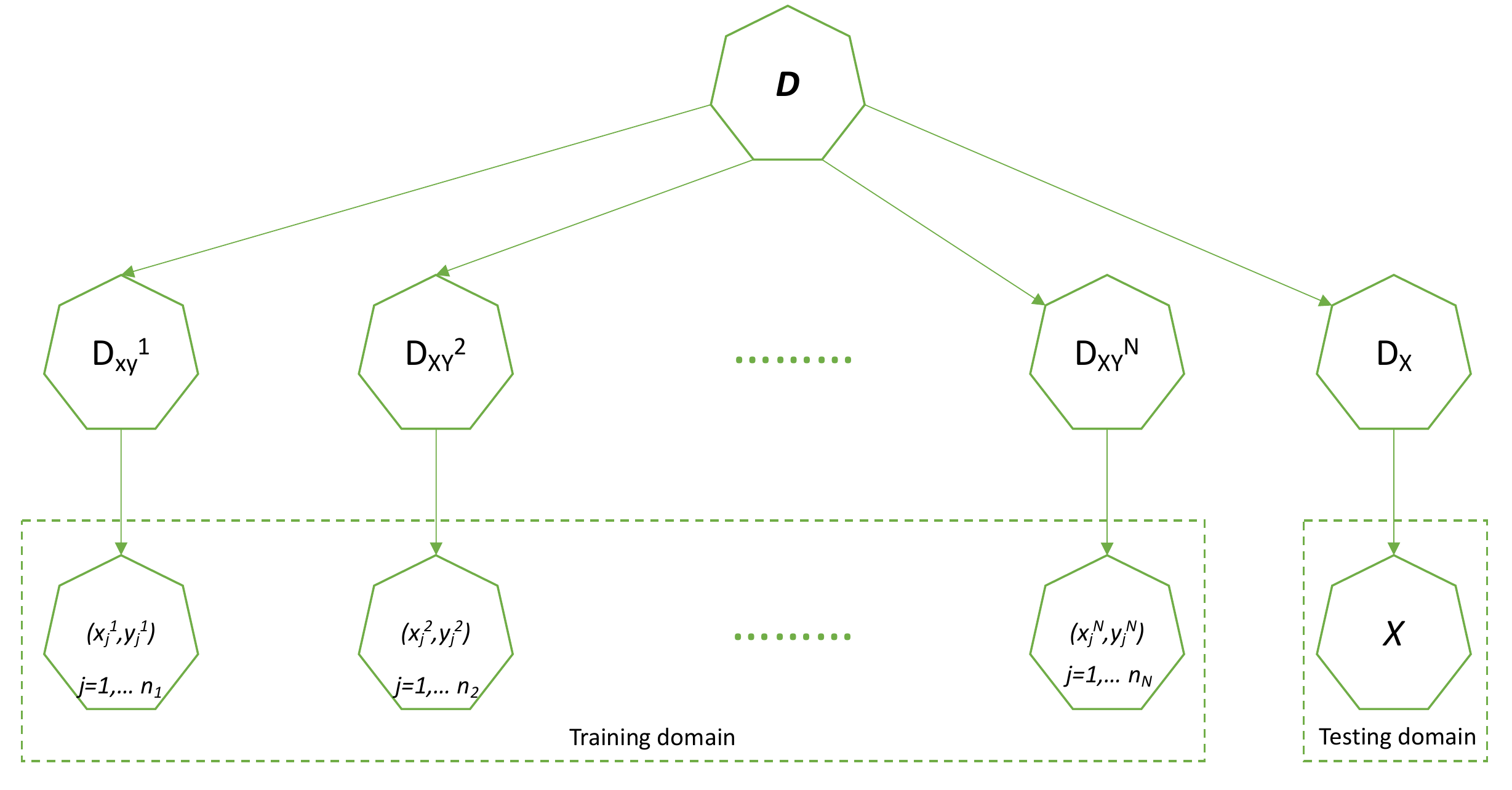}%
\caption{A graphical representation of domain generalisation}
\label{fig3}
\end{figure}

In domain generalisation, let us assume we are given  $\mathcal{N}$  training (source) domains, $S_{train} = \{S^i | i=1,...,N\}$ where $S^i=\{x_j^i,y_j^i\}$ denotes the i-th domains. The joint distributions between each domain are different with $D_{XY}^i \neq  D_{XY}^j$, $N \geq j\neq i \geq 1$. The objective of domain generalisation is to learn a robust and comparatively generalised predictive function $f : X \rightarrow Y$ by using N training domains to get minimum error on an unseen test domain $D_X \rightarrow S_{test}$ where $S_{test}$ cannot be accessed in training and $D_{XY}^{test} \neq D_{XY}^i$.
Therefore, the model's goal is to minimise the loss function $L$ on $S_{test}$
\[\min_{f} E_{(x,y)} \in S_{test}[L(f(x),y)] \] 
where $E$ is the expectation and $y$ are the labels. Figure~\ref{fig3} presents a graphical representation of domain generalisation. 

\subsection{Domain generalisation model experiments}

Here we look at the model pipeline for the training and the inference domain generalisation. The
current version of our approach has been executed on two benchmarks PACS and Office-Home and for each of these,  9 models which were commonly used in  recent literature have been used, these models being  GroupDRO, ANDMask, Mixup, MMD, DANN, CORAL, VREx, RSC, and ERM. Experiments were performed using Pytorch as a backend with an Nvidia GPU RTX 3090, with 24 GB memory. 

Experiments were conducted with original images from the benchmarks without any additional data augmentation and in the pre-processing phase only re-sizing of the images was implemented. Each model has its own set of hyperparameters but the common parameters are batch size 32, epochs 120, momentum 0.9, learning rate 0.01, weight decay 0.0005, input size (3, 224, 224), and baseline model Resnet-18. From each source domain of each of the two datasets, PACS and Office-Home, models utilise 80\%\ of the data in training and validation, and keep 20\%\ of the data as the unseen or target domain.

\subsection{Domain-specific (DL) model experiments}
Our domain-specific pipeline has different settings to the domain generalisation pipeline and it also supports the same two benchmarks as well as 7 domain-specific models namely  AlexNet, VGGNet16, ResNet18, ResNet50, InceptionV3, DenseNet121, and SqueezeNet. In this system, we use the same Pytorch environment with the same Nvidia GPU RTX 3090, with 24 GB memory. These models use a fine tuning technique in which pre-trained weights can be used as a feature extractor and the last fully-connected layers could be re-initialised and trained. 

For these experiments, a model does training and validation in one domain and then performs inference in another target domain. For example, in the case of PACS, models explore the domain of ``art painting'' (with 1,638 samples) in training and in validation, and then use 20\%\ of the target domain's ``cartoon images'' (with 410 samples). Similarly, for the Office-Home dataset, models use the domain of ``clipart'' (with 3,492 samples) as the source, and then images categorised as ``real world'' (with 873 samples) as the target. 

During  training, models do not have any access to the target domain and initially models use only one domain as a source. The common hyperparamters are batch size 64, epochs 120 with early stopping 20, momentum 0.9, learning rate 0.0001, weight decay 0.0005, input size (3, 224, 224), and cross-entropy loss.
\section{Results}
%\Reviewer{Reviewer 1 wants us to change the term from training models to training frameworks for 9 models of domain generalisation}
Table~\ref{tab3} presents the results for 9   domain generalisation frameworks and 7 conventional deep learning or domain-specific models shown in blue font.

\begin{table}[ht]
%\caption{Comparison of classification results for domain generalisation and domain- specific methods} % title of Table
\caption{Experiments with domain generalisation and domain-specific methods} % title of Table
%\Reviewer{Reviewer 1 comment on table's caption}
\centering % used for centering table
%\small
\begin{tabular}{lcccl}\toprule& \multicolumn{2}{c}{PACS} & \multicolumn{2}{c}{Office-Home}
\\%\cmidrule(lr){2-3}\cmidrule(lr){4-5}
\midrule
Models           & ~~~Validation~~~  & ~~~Target~~~    & ~~~Validation~~~  & ~~~Target~~~ \\\midrule
GroupDRO    & 0.95 & 0.73 & 0.82 & 0.52   \\
ANDMask & 0.95 & 0.72 &0.81 & 0.44   \\
Mixup & 0.97 & 0.72 & 0.83 & 0.53  \\
MMD   & 0.94 & 0.69 & 0.82 & 0.52  \\
DANN   & 0.94 & 0.73 & 0.83 & 0.51  \\
CORAL   & 0.95 & 0.77 & 0.84 & 0.55  \\
VREx   & 0.97 & 0.80 & 0.76 & 0.49  \\
RSC   & 0.97 & 0.77 & 0.83 & 0.50  \\
ERM   & 0.97 & 0.78 & 0.84 & 0.57  \\
\midrule % inserts single horizontal line
\color{blue}AlexNet   & 0.74 & 0.45 & 0.56 & 0.30  \\
\color{blue}VGGNet16   & 0.80 & 0.47 & 0.50 & 0.23  \\
\color{blue}ResNet18   & 0.86 & 0.51 & 0.65 & 0.52  \\
\color{blue}ResNet50   & 0.89 & 0.57 & 0.70 & 0.62  \\
\color{blue}InceptionV3   & 0.90 & 0.55 & 0.68 & 0.66 \\
\color{blue}DenseNet121   & 0.86 & 0.44 & 0.62 & 0.35 \\
\color{blue}SqueezeNet   & 0.80 & 0.50 & 0.54 & 0.29 \\\bottomrule
\label{tab3} % is used to refer this table in the text
\end{tabular}
\end{table}

In  Table~\ref{tab3}, the columns marked ``Validation'' and ``Target'' represent the accuracy figures for the validation and for the unseen or target testing set respectively with results presented for  two different benchmarks. Well-trained models will have adequately high validation accuracy and target accuracy always tries to follow validation accuracy. According to various types of data distribution, a model can have different values of validation and target accuracy but for a balanced dataset, an accuracy figure close to 90\%\  can be considered good enough to deploy in some application domains.

Table~\ref{tab3} has five columns and the rows include the names of the models used and the validation and target accuracy performance figures for both datasets. The first 9 models belong to the domain generalisation method and the remaining 7 models in blue  relate to domain-specific methods.
\begin{figure*}[htb!]
\includegraphics[width=\textwidth]{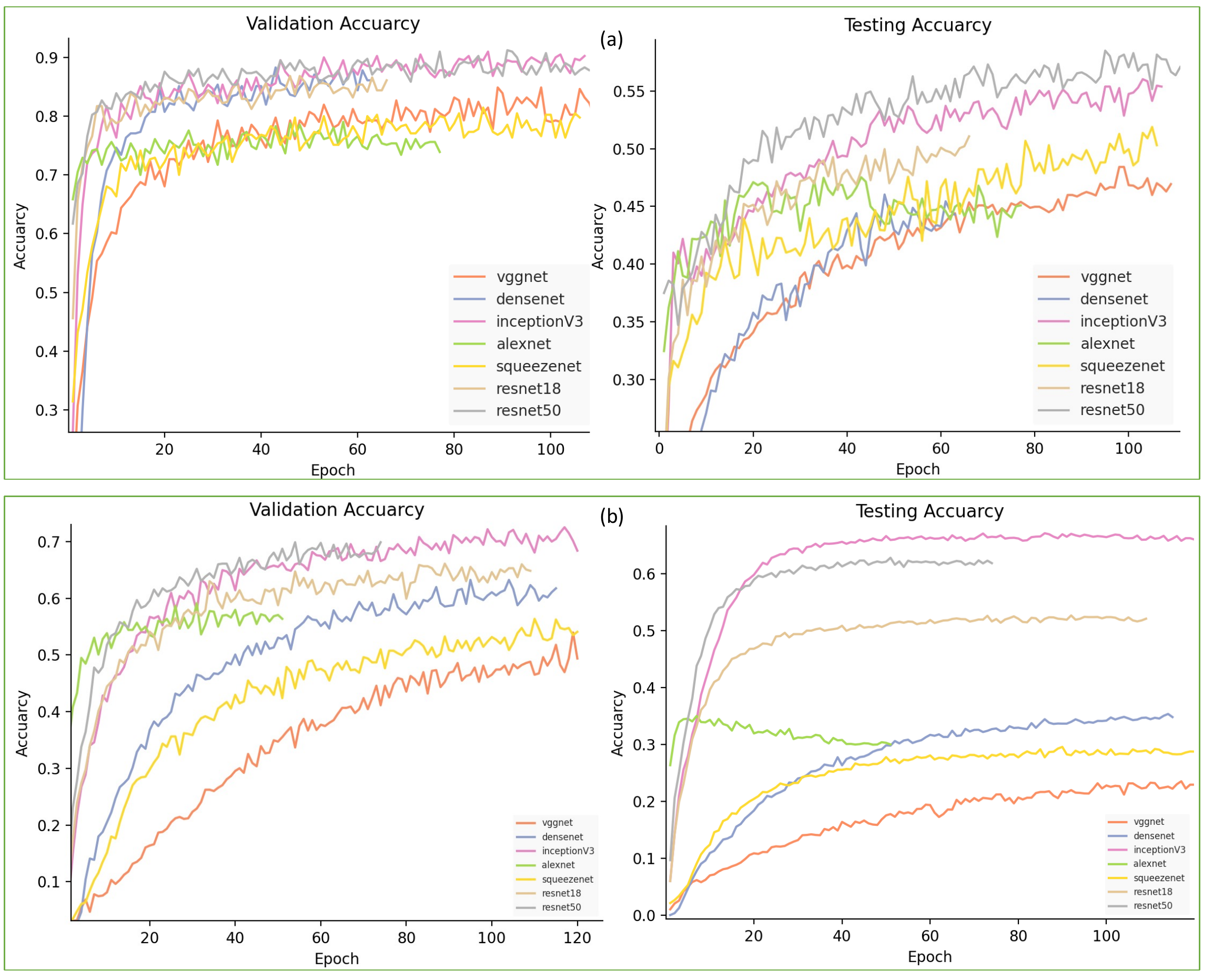} %
\caption{Accuracy analysis for the PACS  and the Office-Home  benchmarks. The length of the curves indicate the stopping points.}
\label{fig4}
\end{figure*}
Overall, if we examine the PACS dataset, domain generalisation methods have clearly higher validation and target accuracy compared to domain-specific methods and VREx  shows the best performance compared to the others. In the case of domain-specific models, InceptionV3 performs better than the others. Similarly, for the Office-Home dataset, both types of models have identical behaviour. 

In the case of domain-specific models, they perform close to the performance levels of the domain generalisation methods for the PACS dataset but we need to consider that during the training, they use a single domain and are tested on another single domain and results on the remaining domains will be different. 

On the other hand, the Office-Home benchmark has more complex tasks than PACS therefore domain-specific models perform comparatively  poorly. From this we can conclude that with more  variation in tasks and domains, the performance of conventional deep learning methods is not stable and fluctuates rapidly.

Even though we try to explain domain generalisation with the help of accuracy matrices, there is no absolute way to measure the performance of domain generalisation. For example, sometimes, a model can have low scores during the training and validation but that model can still have better generalisation because the model will be more stable towards unseen data domains. 

We now present a more detailed analysis of accuracy and loss for both benchmarks. Figure~\ref{fig4} represents the validation and testing accuracy across the datasets. Figure~\ref{fig4}(a) highlights the accuracy curves for PACS and Figure~\ref{fig4}(b) presents accuracy for Office-Home. The graph shows information for 7   conventional deep learning models. In Figure~\ref{fig4}(a) Alexnet shows the lowest accuracy, Resnet50 and InceptionV3 show the highest but almost the same accuracy level, which is around 90\%\ for the validation case. Moreover, in the case of testing/unseen accuracy, Alexnet and VGGNet have almost the same but lowest accuracy from among the others and Resnet50 clearly outperforms other models. From Figure~\ref{fig4}(a), for the PACS benchmark, the key information which we can extract is that less deep or smaller models show low accuracy relative to larger models that give high accuracy. Therefore, we can also say that larger models have better generalisation properties for out-of-distribution datasets.

Figure~\ref{fig4}(b) illustrates accuracy analysis for the Office-Home benchmark dataset and in the case of validation curves, InceptionV3 and Resnet50  show the highest results and VGGNet  performs poorly. On the other hand, if we look at  Table~\ref{tab3}, even though Resnet50 has 70\%\ and InceptionV3 68\%\ validation accuracy and based upon these numbers we can not possibly say that overall Resnet50 has better generalisation. The reason behind this argument is again numbers in table~\ref{tab3} for testing or target set which means that on the target or unseen domains it is InceptionV3 that has higher accuracy among all. Figure~\ref{fig4}(b) testing accuracy curves also contain similar trends like validations curves but one vital piece of information which we can clearly see in it, is the gaps between InceptionV3 and Resnet50 increased. Therefore, for Office-Home, InceptionV3 has better domain generalisation than the other models.

\begin{figure*}[htb!]
\includegraphics[width=\textwidth]{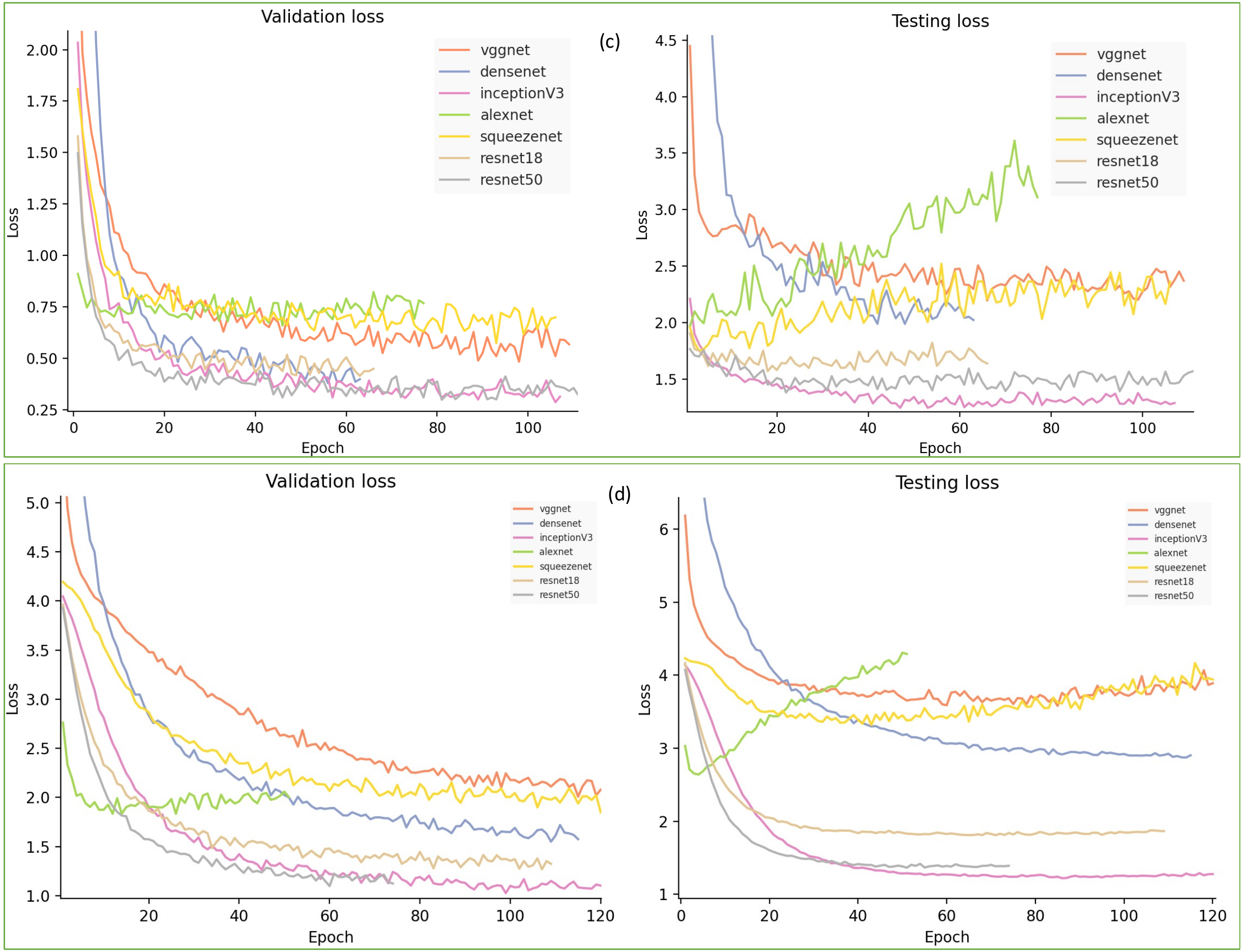}
\caption{Loss analysis of conventional models for the PACS (top) and the Office-Home (bottom) benchmarks.}
\label{fig5}
\end{figure*}

Figure~\ref{fig5}(c) is the validation and testing losses for the PACS benchmark and Alexnet performs worst in both cases and InceptionV3 perform best in both cases. Similar to the accuracy pattern in Office-Home, Figure~\ref{fig5}(c) also shows increments in gaps between the losses of Resnet50 and InceptionV3. Figure~\ref{fig5}(d) illustrates losses for Office-Home. Likewise the accuracy analysis of Office-Home, for the validation loss, VGGNet  give high losses, Resnet50 and InceptionV3 are close to each other but have lowest losses. On unseen data, overall VGGNet shows relatively stable behaviour and Alexnet crosses VGGNet around 40 epochs and becomes a higher loss-giving network. Correspondingly, Resnet50 performs well at the start of analysis but around 35 epochs InceptionV3  crosses Resnet50 and becomes the lowest loss-giving network. Hence, based on losses curves, InceptionV3 has lower losses than other networks and it supports our above-mentioned hypothesis that InceptionV3 has better domain generalisation ability than other conventional models. 
%\Reviewer{Reviewer2: Before concluding, give a discussion that summarizes the data and facts obtained to support the purpose of writing this paper}
\section{Discussion}
The main purpose of this article is related to performance analysis for popular benchmarks of domain generalisation. It also contains experiments for conventional domain specific deep learning methods and recent domain generalisation training frameworks. The results section especially Table~\ref{tab3}, Figure~\ref{fig4} and Figure~\ref{fig5} convey the vital message that domain specific models perform poorly most of the time if we try to explain generalisation with accuracy and loss matrices. Moreover, this article tries to highlight another parameter using graphs in Figure~\ref{fig4} and Figure~\ref{fig5} which show the gaps between validation and testing accuracy. Higher gaps mean the target model has poor domain generalisation and lower gaps mean comparatively higher generalisation.

Another outcome which we can extract from the findings of this article is that even domain specific models perform less effectively when we compare them with domain generalisation frameworks but larger models have better domain generalisation. Therefore, in Table~\ref{tab3} ResNet50 has the best domain generalisation results for both benchmarks including PACS and Office-Home compared to other domain specific models in blue colour. Furthermore, models having skip connection like ResNets, DenseNets are better for domain generalisation compared to models without skip connections like AlexNet and VGGNets.

%\Reviewer{Reviewer2: Future plans are not discussed well. What questions remain in the work, any limitation to be fixed, any improvement?}
%\Reviewer{Reviewer2: It will be nice if the author(s) indicate whether or not the study has achieved its aim and objectives at the end of the paper.}
\section{Conclusions and Future Plans}
This paper presents important information in the form of a summary of the performances of various vision-based machine learning tools and  connects these results with the emerging areas of domain generalisation and domain adaptation. The two base pipelines presented help us to understand that for the PACS and Office-Home benchmarks,  domain-specific methods perform poorly. 

The  work here successfully demonstrates that in the field of supervised learning, domain generalisation learning is better than domain-specific learning for some kinds of benchmarks. Our evaluation is performed  on relatively complex benchmarks and by determining their accuracy, we try to explain generalisation. 

Other ways for measuring  domain generalisation have been proposed in the literature like measuring the gap between source and target domains, which is one of our future research directions. Furthermore, we will extend our experiments to cover attention based vision transformers as it would be insightful to introduce an attention mechanism for such benchmarks. Meanwhile our present results are for benchmarks which have 4 domains, and as next steps we will increase the number of domains by using benchmarks like DomainNet which has 345 classes. To test domain generalisation for OOD we will create our own testing benchmark.

This article has achieved its aims partially as we were able to test the findings for only supervised learning with vision based benchmarks. Hence, it would be interesting to explore other areas and applications like unsupervised learning and audio benchmarks. Besides such concerns, our work highlights important future research directions to explore domain generalisation.

\section*{Acknowledgments}
HR and this publication has emanated from research conducted with the financial
support of Science Foundation Ireland under Grant number 18/CRT/6183. For the purpose
of Open Access, the author has applied a CC-BY public copyright licence to any
author accepted Manuscript version arising from this submission.
AS is part-funded by Science Foundation Ireland (SFI) under Grant Number SFI/12/RC/2289\_P2 (Insight SFI Research Centre for Data Analytics), co-funded by the European Regional Development Fund.


\vspace{12pt}
%\bibliography{paper}
\end{document}